\documentclass{article} 

\usepackage{nips13submit_e,times}
\usepackage{graphicx}
\usepackage{hyperref}
\usepackage{url}
\usepackage{natbib}
\usepackage{amsmath}
\usepackage{amssymb}
\usepackage{subfigure}
\usepackage{color}
\usepackage{caption}
\usepackage{authblk}
\usepackage{natbib}
\usepackage{multirow}
\usepackage{wrapfig}

\newcommand{\bmx}[0]{\begin{bmatrix}}
\newcommand{\emx}[0]{\end{bmatrix}}

\newcommand{\vect}[1]{\mathbf{#1}}
\newcommand{\vects}[1]{\boldsymbol{#1}}
\newcommand{\matr}[1]{\mathbf{#1}}

\newcommand{\va}[0]{\vect{a}}
\newcommand{\vb}[0]{\vect{b}}

\newcommand{\vh}[0]{\vect{h}}

\newcommand{\vx}[0]{\vect{x}}

\newcommand{\vy}[0]{\vect{y}}

\newcommand{\mW}[0]{\matr{W}}

\newcommand{\mU}[0]{\matr{U}}
\newcommand{\mV}[0]{\matr{V}}

\newcommand{\vzero}[0]{\vect{0}}

\newcommand{\TT}[0]{{\vects{\theta}}}

\newcommand{\specialcell}[2][c]{%
  \begin{tabular}[#1]{@{}c@{}}#2\end{tabular}}

\nipsfinalcopy 

\title{How to Construct Deep Recurrent Neural Networks}

\author[1]{Razvan Pascanu}
\author[1]{Caglar Gulcehre}
\author[2]{Kyunghyun Cho}
\author[1]{Yoshua Bengio}

\affil[1]{D\'{e}partement d'Informatique et de Recherche Op\'{e}rationelle\\
Universit\'{e} de Montr\'{e}al\\
    \texttt{\{pascanur, gulcehrc\}@iro.umontreal.ca, yoshua.bengio@umontreal.ca}}
\affil[2]{Department of Information and Computer Science\\ Aalto University School of Science\\
    \texttt{kyunghyun.cho@aalto.fi}}

\begin{document}

\maketitle

\begin{abstract}
    In this paper, we explore different ways to extend a recurrent
    neural network (RNN) to a \textit{deep} RNN. We start by
    arguing that the concept of depth in an RNN is not as clear
    as it is in feedforward neural networks.  By carefully
    analyzing and understanding the architecture of an RNN,
    however, we find three points of an RNN which may be made
    deeper; (1) input-to-hidden function, (2) hidden-to-hidden
    transition and (3) hidden-to-output function. Based on this
    observation, we propose two novel architectures of a deep RNN
    which are orthogonal to an earlier attempt of stacking
    multiple recurrent layers to build a deep RNN
    \citep{Schmidhuber1992,ElHihi+Bengio-nips8}.  We provide an alternative
    interpretation of these deep RNNs using a novel framework
    based on neural operators. The proposed deep RNNs are
    empirically evaluated on the tasks of polyphonic music
    prediction and language modeling. The experimental result
    supports our claim that the proposed deep RNNs benefit from
    the depth and outperform the conventional, shallow RNNs.
\end{abstract}

\section{Introduction}

Recurrent neural networks~\citep[RNN, see, e.g.,][]{Rumelhart86b} have recently
become a popular choice for modeling variable-length sequences.  RNNs have been
successfully used for various task such as language modeling~\citep[see,
e.g.,][]{Graves2013,Pascanu+al-ICML2013-small,Mikolov-thesis-2012,Sutskever2011},
learning word embeddings~\citep[see, e.g.,][]{Mikolov2013}, online handwritten
recognition~\citep{Graves09} and speech recognition~\citep{Graves13speech}.

In this work, we explore \textit{deep} extensions of the basic RNN. Depth for
feedforward models can lead to more expressive models~\citep{Pascanu2014}, and
we believe the same should hold for recurrent models.  We claim that, unlike in
the case of feedforward neural networks, the \textit{depth} of an RNN is
ambiguous.  In one sense, if we consider the existence of a composition of
several nonlinear computational layers in a neural network being deep, RNNs are
already deep, since any RNN can be expressed as a composition of multiple
nonlinear layers when unfolded in time.  

\citet{Schmidhuber1992, ElHihi+Bengio-nips8} earlier proposed another way of building a deep RNN by
stacking multiple recurrent hidden states on top of each other. This approach
potentially allows the hidden state at each level to operate at different
timescale~\citep[see, e.g.,][]{Hermans2013}.  Nonetheless, we notice that there
are some other aspects of the model that may still be considered
\textit{shallow}. For instance, the transition between two consecutive hidden
states at a single level is shallow, when viewed separately.This has
implications on what kind of transitions this model can represent as discussed
in Section~\ref{sec:deep_hid_to_hid}.

Based on this observation, in this paper, we investigate possible approaches to
extending an RNN into a deep RNN. We begin by studying which parts of an RNN
may be considered shallow.  Then, for each shallow part, we propose an
alternative \textit{deeper} design, which leads to a number of deeper variants
of an RNN. The proposed deeper variants are then empirically evaluated on two
sequence modeling tasks.

The layout of the paper is as follows.  In Section~\ref{sec:rnn} we briefly
introduce the concept of an RNN. In Section~\ref{sec:drnn} we explore different
concepts of \textit{depth} in RNNs.  In particular, in
Section~\ref{sec:dt_rnn}--\ref{sec:dot_rnn} we propose two novel variants of
deep RNNs and evaluate them empirically in Section~\ref{sec:experiments} on two
tasks: polyphonic music prediction~\citep{Boulanger+al-ICML2012-small} and
language modeling.  Finally we discuss the shortcomings and advantages of the
proposed models in Section~\ref{sec:final}.

\section{Recurrent Neural Networks}
\label{sec:rnn}

A recurrent neural network (RNN) is a neural network that simulates a
discrete-time dynamical system that has an input $\vx_t$, an output $\vy_t$ and
a hidden state $\vh_t$.  In our notation the subscript $t$ represents time.
The dynamical system is defined by
\begin{align}
    \label{eq:dynamical_system_trans}
    \vh_t &= f_h(\vx_t, \vh_{t-1}) 
    \\
    \label{eq:dynamical_system_out}
    \vy_t &= f_o(\vh_t),
\end{align}
where $f_h$ and $f_o$ are a state transition function and an
output function, respectively. Each function is parameterized by
a set of parameters; $\TT_h$ and $\TT_o$.

Given a set of $N$ training sequences 
$D=\left\{ \left(
    (\vx_1^{(n)}, \vy_1^{(n)} ), \dots, (\vx_{T_n}^{(n)},
    \vy_{T_n}^{(n)}) \right)\right\}_{n=1}^N$, 
the parameters of an RNN can be estimated by minimizing the
following cost function:
\begin{align}
    \label{eq:cost_function}
    J(\TT) = \frac{1}{N} \sum_{n=1}^N \sum_{t=1}^{T_n} 
    d(\vy_t^{(n)}, f_o(\vh_t^{(n)})),
\end{align}
where $\vh_t^{(n)}=f_h(\vx_t^{(n)}, \vh_{t-1}^{(n)})$ and
$\vh_0^{(n)}=\vzero$. $d(\va, \vb)$ is a predefined divergence
measure between $\va$ and $\vb$, such as Euclidean distance or
cross-entropy.

\subsection{Conventional Recurrent Neural Networks}

A conventional RNN is constructed by defining the transition
function and the output function as
\begin{align}
    \label{eq:rnn_shallow_trans}
    \vh_t &= f_h(\vx_t, \vh_{t-1}) = \phi_h\left(\mW^\top
    \vh_{t-1} + \mU^\top \vx_t \right) \\
    \label{eq:rnn_shallow_out}
    \vy_t &= f_o(\vh_t, \vx_t) = \phi_o\left(\mV^\top
    \vh_{t}\right),
\end{align}
where $\mW$, $\mU$ and $\mV$ are respectively the transition, input and output
matrices, and $\phi_h$ and $\phi_o$ are element-wise nonlinear functions. It is
usual to use a saturating nonlinear function such as a logistic sigmoid
function or a hyperbolic tangent function for $\phi_h$. An illustration of this
RNN is in Fig.~\ref{fig:rnn_models} (a).

The parameters of the conventional RNN can be estimated by, for instance,
stochastic gradient descent (SGD) algorithm with the gradient of the cost
function in Eq.~\eqref{eq:cost_function} computed by backpropagation through
time~\citep{Rumelhart86b}.

\section{Deep Recurrent Neural Networks}
\label{sec:drnn}

\subsection{Why Deep Recurrent Neural Networks?}

Deep learning is built around a hypothesis that a deep,
hierarchical model can be exponentially more efficient at
representing some functions than a shallow one~\citep{Bengio2009FTML}. A number of recent theoretical results
support this hypothesis~\citep[see,
e.g.,][]{Roux2010,Delalleau+Bengio-2011-small, Pascanu2014}. For instance, it
has been shown by \citet{Delalleau+Bengio-2011-small} that a deep
sum-product network may require exponentially less units to
represent the same function compared to a shallow sum-product
network. Furthermore, there is a wealth of empirical evidences
supporting this hypothesis~\citep[see,
e.g.,][]{Goodfellow_maxout_2013,Hinton-et-al-arxiv2012,Hinton-et-al-2012}.
These findings make us suspect that the same argument should
apply to recurrent neural networks. 

\subsection{Depth of a Recurrent Neural Network}
\label{sec:depth_rnn}

\begin{figure}[ht]
    \centering
    \includegraphics[width=0.4\columnwidth]{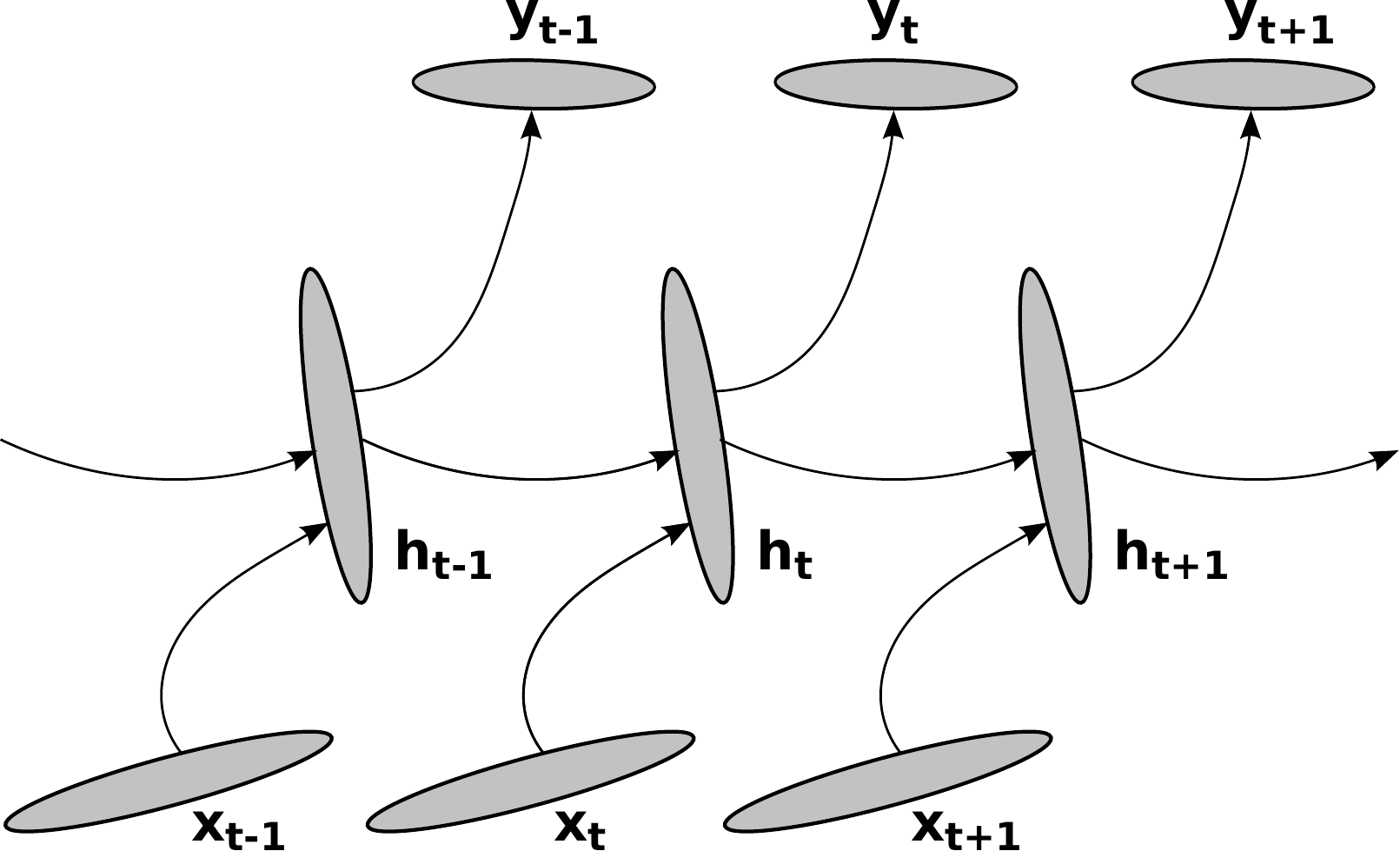}
\caption{A conventional recurrent neural network unfolded in time.}
\label{fig:model1_unfolded}
\vskip -5mm
\end{figure}

The \textit{depth} is defined in the case of feedforward neural networks as
having multiple nonlinear layers between input and output. Unfortunately this
definition does not apply trivially to a recurrent neural network (RNN) because
of its temporal structure. For instance, any RNN when unfolded in time as in
Fig.~\ref{fig:model1_unfolded} is deep, because a computational path between
the input at time $k < t$ to the output at time $t$ crosses several nonlinear
layers. 

A close analysis of the computation carried out by an RNN (see
Fig.~\ref{fig:rnn_models} (a)) at each time step individually, however, shows
that certain transitions are not deep, but are only results of a linear
projection followed by an element-wise nonlinearity.  It is clear that the
hidden-to-hidden ($\vh_{t-1} \to \vh_{t}$), hidden-to-output ($\vh_{t} \to
\vy_{t}$) and input-to-hidden ($\vx_t \to \vh_t$) functions are all
\textit{shallow} in the sense that there exists no intermediate, nonlinear
hidden layer. 

We can now consider different types of depth of an RNN by considering those
transitions separately.  We may make the hidden-to-hidden transition deeper by
having one or more intermediate nonlinear layers between two consecutive hidden
states ($\vh_{t-1}$ and $\vh_{t}$). At the same time, the hidden-to-output
function can be made deeper, as described previously, by plugging, multiple
intermediate nonlinear layers between the hidden state $\vh_t$ and the output
$\vy_t$.  Each of these choices has a different implication. 

\subsubsection{Deep Input-to-Hidden Function}

A model can exploit more non-temporal structure from the input by making the
input-to-hidden function deep.  Previous work has shown that higher-level
representations of deep networks tend to better disentangle the underlying
factors of variation than the original
input~\citep{Goodfellow2009,Glorot+al-ICML-2011-small} and flatten the
manifolds near which the data concentrate~\citep{Bengio-et-al-ICML2013}.  We
hypothesize that such higher-level representations should make it easier to
learn the temporal structure between successive time steps because the
relationship between abstract features can generally be expressed more easily.
This has been, for instance, illustrated by the recent
work~\citep{Mikolov-et-al-ICLR2013} showing that word embeddings from neural
language models tend to be related to their temporal neighbors by simple
algebraic relationships, with the same type of relationship (adding a vector)
holding over very different regions of the space, allowing a form of analogical
reasoning. 

This approach of making the input-to-hidden function deeper is in the line with
the standard practice of replacing input with extracted features in order to
improve the performance of a machine learning model~\citep[see,
e.g.,][]{Bengio2009FTML}.  Recently, \citet{Chen2013} reported that a better
speech recognition performance could be achieved by employing this strategy,
although they did not jointly train the deep input-to-hidden function together
with other parameters of an RNN. 

\subsubsection{Deep Hidden-to-Output Function}

A deep hidden-to-output function can be useful to disentangle the factors of
variations in the hidden state, making it easier to predict the output. This
allows the hidden state of the model to be more compact and may result in the
model being able to summarize the history of previous inputs more efficiently.
Let us denote an RNN with this deep hidden-to-output function a deep output RNN
(DO-RNN). 

Instead of having feedforward, intermediate layers between the hidden state and
the output, \citet{Boulanger+al-ICML2012-small} proposed to replace the output
layer with a conditional generative model such as restricted Boltzmann machines
or neural autoregressive distribution estimator~\citep{Larochelle+Murray-2011}.
In this paper we only consider feedforward intermediate layers.

\subsubsection{Deep Hidden-to-Hidden Transition}
\label{sec:deep_hid_to_hid}

The third knob we can play with is the depth of the hidden-to-hidden
transition.  The state transition between the consecutive hidden states
effectively adds a new input to the summary of the previous inputs represented
by the fixed-length hidden state. Previous work with RNNs has generally limited
the architecture to a shallow operation; affine transformation followed by an
element-wise nonlinearity. Instead, we argue that this procedure of
constructing a new summary, or a hidden state, from the combination of the
previous one and the new input should be highly nonlinear. This nonlinear
transition could allow, for instance, the hidden state of an RNN to rapidly
adapt to quickly changing modes of the input, while still preserving a useful
summary of the past. This may be impossible to be modeled by a function from
the family of generalized linear models. However, this highly nonlinear
transition can be modeled by an MLP with one or more hidden layers which has an
universal approximator property~\citep[see, e.g.,][]{Hornik89}.

\begin{figure}[t]
    \centering
    \begin{minipage}{0.19\textwidth}
        \centering
        \includegraphics[width=0.95\columnwidth]{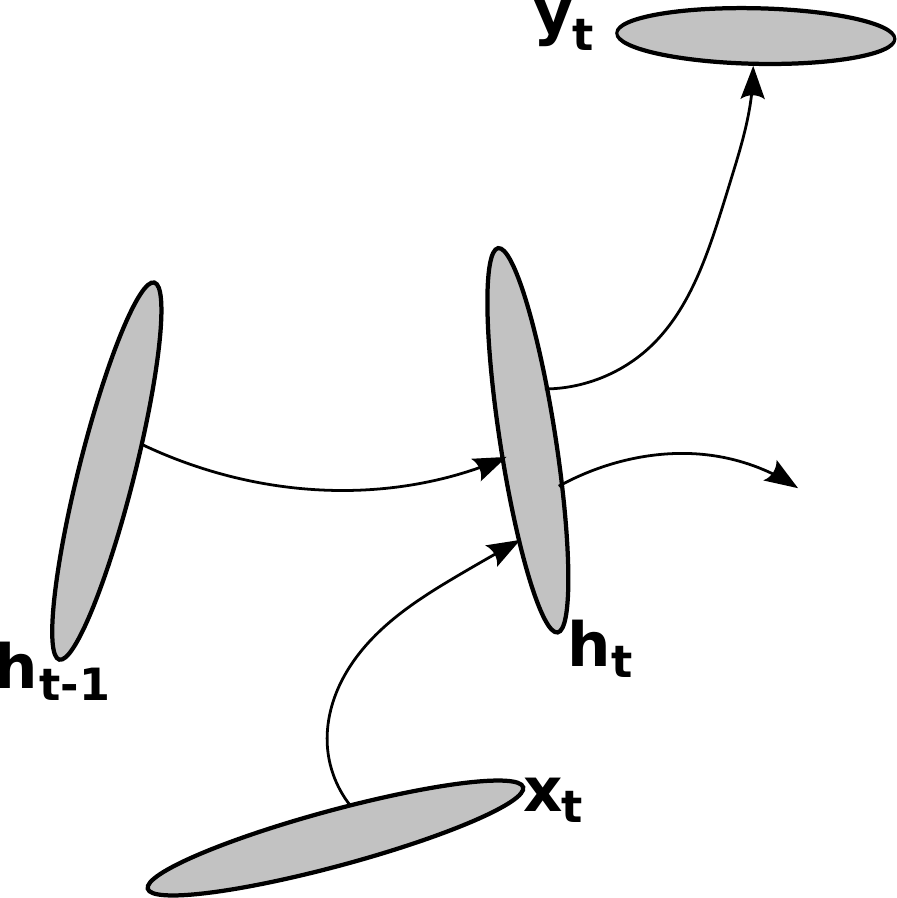}
    \end{minipage}
    \hfill
    \begin{minipage}{0.19\textwidth}
        \centering
        \includegraphics[width=0.95\columnwidth]{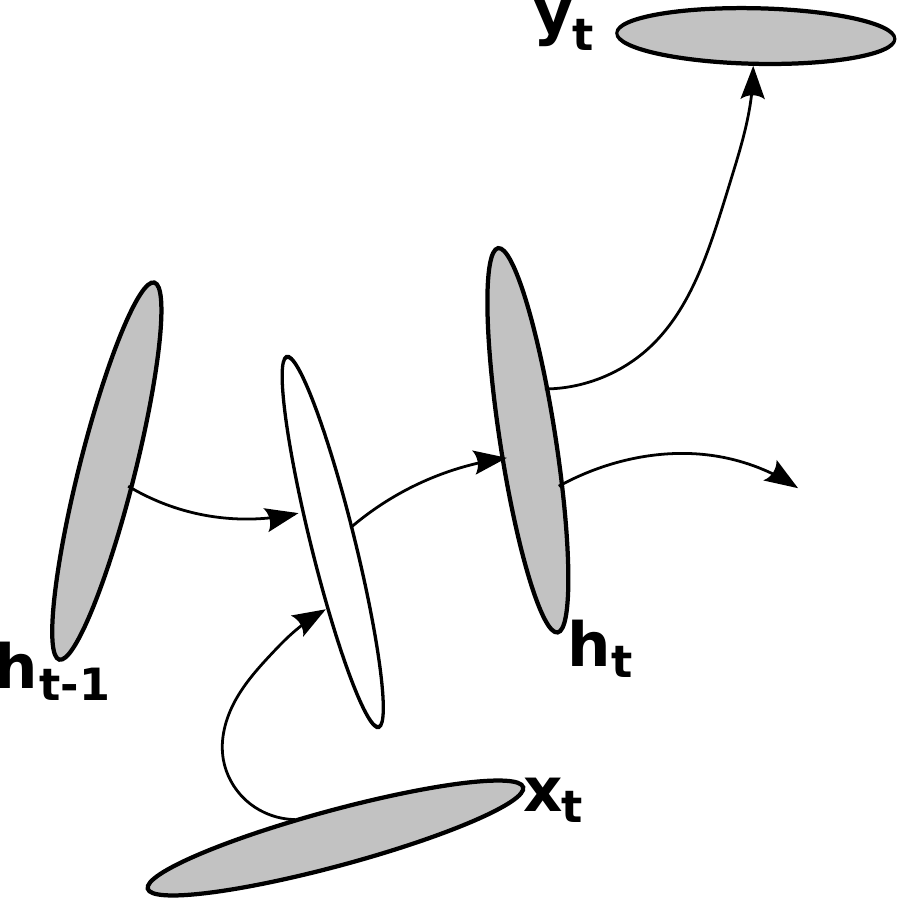}
    \end{minipage}
    \hfill
    \begin{minipage}{0.19\textwidth}
        \centering
        \includegraphics[width=0.95\columnwidth]{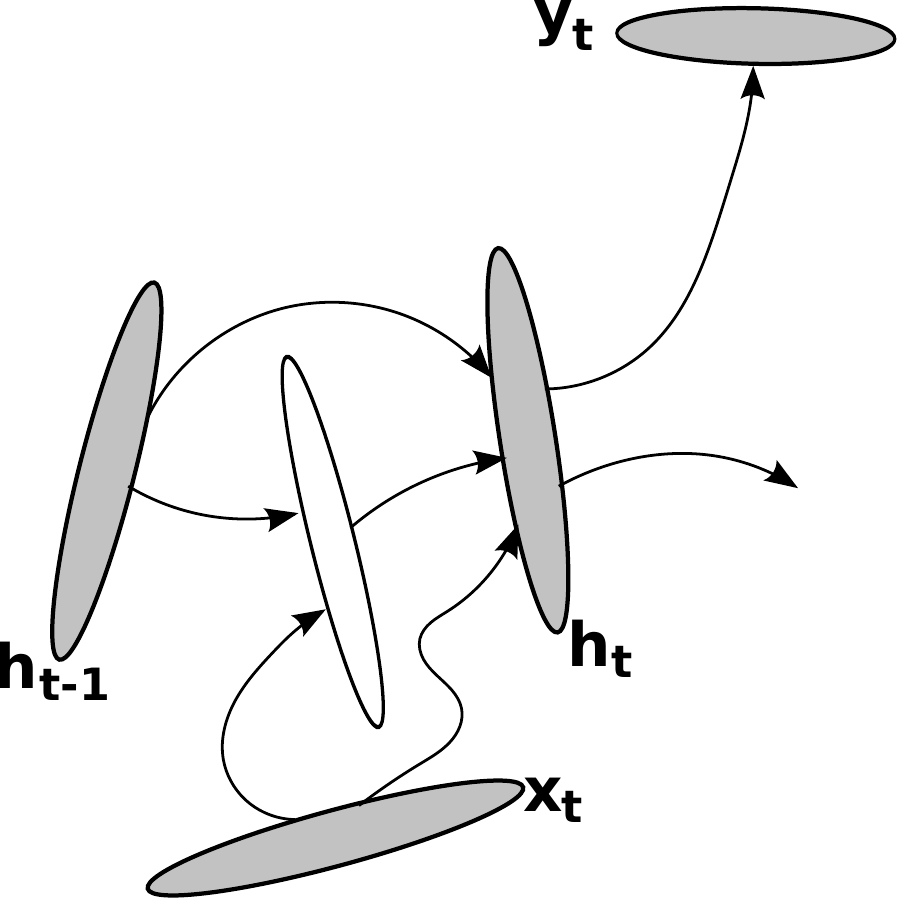}
    \end{minipage}
    \hfill
    \begin{minipage}{0.19\textwidth}
        \centering
        \includegraphics[width=0.95\columnwidth]{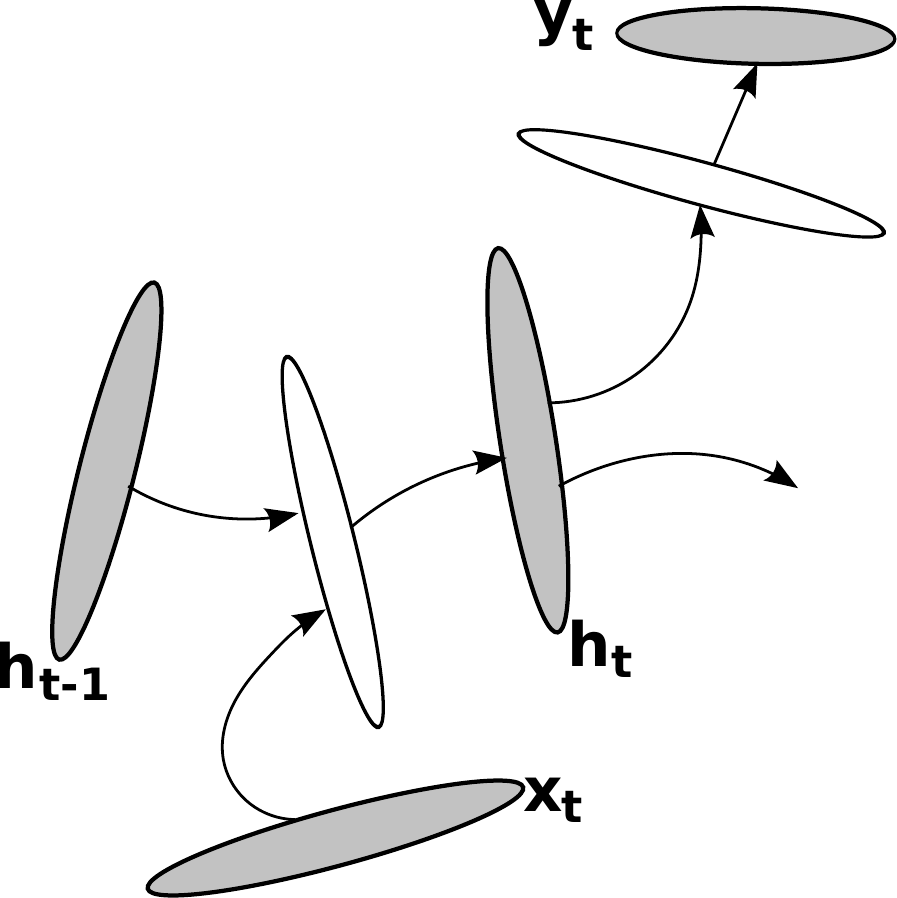}
    \end{minipage}
    \hfill
    \begin{minipage}{0.19\textwidth}
        \centering
        \includegraphics[width=0.95\columnwidth]{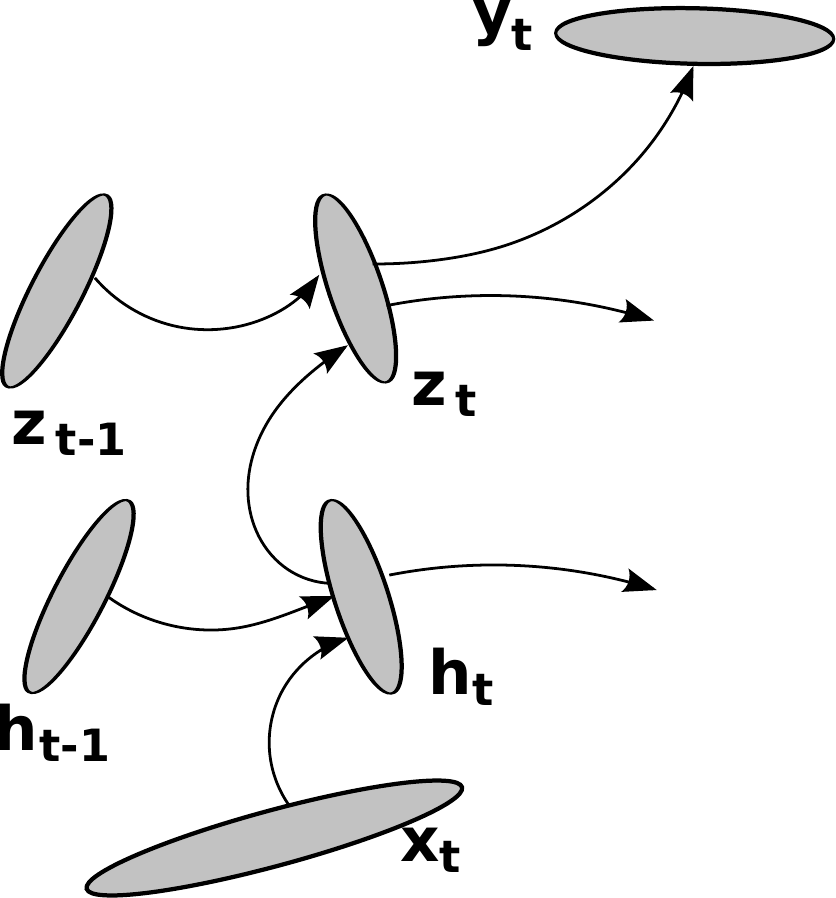}
    \end{minipage}
    \\
    \vspace{3mm}
    \begin{minipage}{0.19\textwidth}
        \centering
        (a) RNN
    \end{minipage}
    \hfill
    \begin{minipage}{0.17\textwidth}
        \centering
        (b) DT-RNN
    \end{minipage}
    \hfill
    \begin{minipage}{0.19\textwidth}
        \centering
        (b*) DT(S)-RNN
    \end{minipage}
    \hfill
    \begin{minipage}{0.19\textwidth}
        \centering
        (c) DOT-RNN
    \end{minipage}
    \hfill
    \begin{minipage}{0.19\textwidth}
        \centering
        (d) Stacked RNN
    \end{minipage}
\caption{Illustrations of four different recurrent neural networks (RNN). (a) A
conventional RNN. (b) Deep Transition (DT) RNN. (b*) DT-RNN with shortcut
connections (c) Deep Transition, Deep Output (DOT) RNN. (d) Stacked RNN}
\label{fig:rnn_models}
\vskip -5mm
\end{figure}

An RNN with this deep transition will be called a deep transition RNN (DT-RNN)
throughout remainder of this paper. This model is shown in
Fig.~\ref{fig:rnn_models} (b). 

This approach of having a deep transition, however, introduces a potential
problem. As the introduction of deep transition increases the number of
nonlinear steps the gradient has to traverse when propagated back in time, it
might become more difficult to train the model to capture long-term
dependencies~\citep{Bengio-trnn93}. One possible way to address this difficulty
is to introduce shortcut connections~\citep[see, e.g.,][]{Raiko2012} in the
deep transition, where the added shortcut connections provide shorter paths,
skipping the intermediate layers, through which the gradient is propagated back
in time.  We refer to an RNN having deep transition with shortcut connections
by DT(S)-RNN (See Fig.~\ref{fig:rnn_models} (b*)). 

Furthermore, we will call an RNN having both a deep hidden-to-output function
and a deep transition a deep output, deep transition RNN (DOT-RNN).  See
Fig.~\ref{fig:rnn_models} (c) for the illustration of DOT-RNN. If we consider
shortcut connections as well in the hidden to hidden transition, we call the
resulting model DOT(S)-RNN.  

An approach similar to the deep hidden-to-hidden transition has been proposed
recently by \citet{Pinheiro2013} in the context of parsing a static scene.  They
introduced a recurrent convolutional neural network (RCNN) which can be
understood as a recurrent network whose the transition between consecutive
hidden states (and input to hidden state) is modeled by a convolutional neural
network. 
The RCNN was shown to speed up scene parsing and obtained the state-of-the-art
result in Stanford Background and SIFT Flow datasets. \citet{Ko09} proposed deep
transitions for Gaussian Process models. Earlier, \citet{Valpola2002} used a
deep neural network to model the state transition in a nonlinear, dynamical
state-space model.

\subsubsection{Stack of Hidden States}

An RNN may be extended deeper in yet another way by stacking multiple recurrent
hidden layers on top of each other~\citep{Schmidhuber1992,ElHihi+Bengio-nips8,Jaeger2007,Graves2013}.  We call this model a stacked
RNN (sRNN) to distinguish it from the other proposed variants.  The goal of a
such model is to encourage each recurrent level to operate at a different
timescale. 

It should be noticed that the DT-RNN and the sRNN extend the conventional,
shallow RNN in different aspects. If we look at each recurrent level of the
sRNN separately, it is easy to see that the transition between the consecutive
hidden states is still shallow. As we have argued above,  this limits the
family of functions it can represent. For example, if the structure of the data
is sufficiently complex, incorporating a new input frame into the summary of
what had been seen up to now might be an arbitrarily complex function. In such
a case we would like to model this function by something that has universal
approximator properties, as an MLP. The model can not rely on the higher layers
to do so, because the higher layers do not feed back into the lower layer.  On
the other hand, the sRNN can deal with multiple time scales in the input
sequence, which is not an obvious feature of the DT-RNN. The DT-RNN and the
sRNN are, however, orthogonal in the sense that it is possible to have both
features of the DT-RNN and the sRNN by stacking multiple levels of DT-RNNs to
build a stacked DT-RNN which we do not explore more in this paper.

\subsection{Formal descriptions of deep RNNs}

Here we give a more formal description on how the deep transition
recurrent neural network (DT-RNN) and the deep output RNN
(DO-RNN) as well as the stacked RNN are implemented.

\subsubsection{Deep Transition RNN}
\label{sec:dt_rnn}

We noticed from the state transition equation of the dynamical
system simulated by RNNs in Eq.~\eqref{eq:dynamical_system_trans}
that there is no restriction on the form of $f_h$. Hence, we
propose here to use a multilayer perceptron to approximate $f_h$
instead. 

In this case, we can implement $f_h$ by $L$ intermediate layers
such that
\begin{align*}
    \vh_t &= f_h(\vx_t, \vh_{t-1}) = \phi_h \left(
    \mW_L^\top \phi_{L-1} \left(
    \mW_{L-1}^\top \phi_{L-2} \left(
    \cdots \phi_{1} \left(
    \mW_{1}^\top \vh_{t-1} + \mU^\top \vx_t
    \right)
    \right)
    \right)
    \right),
\end{align*}
where $\phi_{l}$ and $\mW_{l}$ are the element-wise nonlinear
function and the weight matrix for the $l$-th layer.  This RNN
with a multilayered transition function is a deep transition RNN
(DT-RNN). 

An illustration of building an RNN with the deep state transition
function is shown in Fig.~\ref{fig:rnn_models} (b). In the
illustration the state transition function is implemented with a
neural network with a single intermediate layer.

This formulation allows the RNN to learn a non-trivial, highly
nonlinear transition between the consecutive hidden states.  

\subsubsection{Deep Output RNN}
\label{sec:dot_rnn}

Similarly, we can use a multilayer perceptron with $L$
intermediate layers to model the output function $f_o$ in
Eq.~\eqref{eq:dynamical_system_out} such that
\begin{align*}
    \vy_t &= f_o(\vh_t) = \phi_o \left(
    \mV_L^\top \phi_{L-1} \left(
    \mV_{L-1}^\top \phi_{L-2} \left(
    \cdots \phi_{1} \left(
    \mV_{1}^\top \vh_{t}
    \right)
    \right)
    \right)
    \right),
\end{align*}
where $\phi_{l}$ and $\mV_{l}$ are the element-wise nonlinear
function and the weight matrix for the $l$-th layer.  An RNN
implementing this kind of multilayered output function is a deep
output recurrent neural network (DO-RNN).

Fig.~\ref{fig:rnn_models} (c) draws a deep output, deep
transition RNN (DOT-RNN) implemented using both the deep
transition and the deep output with a single intermediate layer
each. 

\subsubsection{Stacked RNN}

The stacked RNN~\citep{Schmidhuber1992,ElHihi+Bengio-nips8} has multiple levels of
transition functions defined by
\begin{align*}
    \vh_t^{(l)} &= f_h^{(l)}(\vh_{t}^{(l-1)}, \vh_{t-1}^{(l)}) = 
    \phi_h\left(\mW_{l}^\top \vh_{t-1}^{(l)} + 
    \mU_{l}^\top \vh_{t}^{(l-1)} \right),
\end{align*}
where $\vh_t^{(l)}$ is the hidden state of the $l$-th level at
time $t$.  When $l=1$, the state is computed using $\vx_t$
instead of $\vh_{t}^{(l-1)}$. The hidden states of all the levels
are recursively computed from the bottom level $l=1$.

Once the top-level hidden state is computed, the output can be
obtained using the usual formulation in
Eq.~\eqref{eq:rnn_shallow_out}. Alternatively, one may use all
the hidden states to compute the output~\citep{Hermans2013}. Each
hidden state at each level may also be made to depend on the
input as well~\citep{Graves2013}. Both of them can be considered
approaches using shortcut connections discussed earlier.

The illustration of this stacked RNN is in
Fig.~\ref{fig:rnn_models} (d). 

\section{Another Perspective: Neural Operators}

In this section, we briefly introduce a novel approach with which
the already discussed deep transition (DT) and/or deep output
(DO) recurrent neural networks (RNN) may be built. We call this
approach which is based on building an RNN with a set of
predefined neural operators, an operator-based framework.

In the operator-based framework, one first defines a set of
operators of which each is implemented by a multilayer perceptron
(MLP). For instance, a \textit{plus} operator $\oplus$ may be
defined as a function receiving two vectors $\vx$ and $\vh$ and
returning the summary $\vh'$ of them:
\begin{align*}
    \vh' = \vx \oplus \vh,
\end{align*}
where we may constrain that the dimensionality of $\vh$ and
$\vh'$ are identical.  Additionally, we can define another
operator $\rhd$ which \textit{predicts} the most likely output
symbol $\vx'$ given a summary $\vh$, such that
\begin{align*}
    \vx' = \rhd \vh
\end{align*}
It is possible to define many other operators, but in this paper,
we stick to these two operators which are sufficient to express
all the proposed types of RNNs.

\begin{wrapfigure}{I}{0.4\textwidth}
    \centering
    \includegraphics[width=0.25\columnwidth]{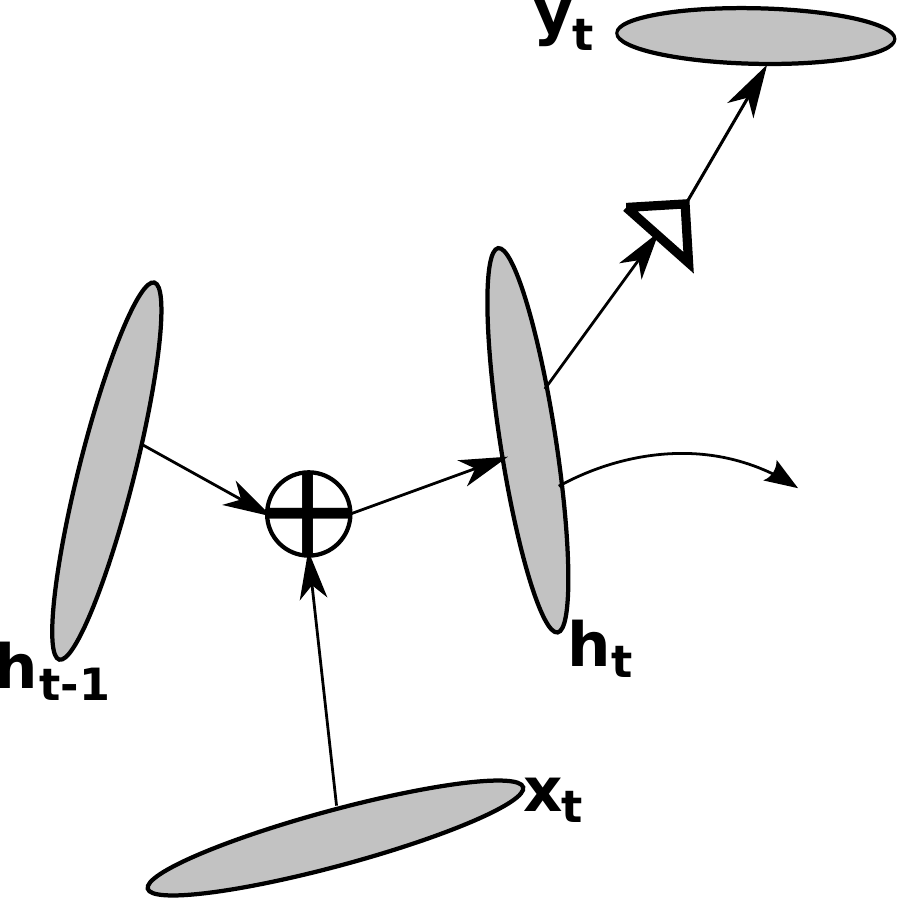}
\caption{A view of
an RNN under the operator-based framework: $\oplus$ and $\rhd$
are the \textit{plus} and \textit{predict} operators,
respectively. }
\label{fig:model_operator}
\vskip -3mm
\end{wrapfigure}

It is clear to see that the plus operator $\oplus$ and the
predict operator $\rhd$ correspond to the transition function and
the output function in
Eqs.~\eqref{eq:dynamical_system_trans}--\eqref{eq:dynamical_system_out}.
Thus, at each step, an RNN can be thought as performing the plus
operator to update the hidden state given an input ($\vh_t =
\vx_t \oplus \vh_{t-1}$) and then the predict operator to compute
the output ($\vy_t = \rhd \vh_t = \rhd (\vx_t \oplus
\vh_{t-1})$). See Fig.~\ref{fig:model_operator} for the
illustration of how an RNN can be understood from the
operator-based framework.

Each operator can be parameterized as an MLP with one or more
hidden layers, hence a neural operator, since we cannot simply
expect the operation will be linear with respect to the input
vector(s).  By using an MLP to implement the operators, the
proposed deep transition, deep output RNN (DOT-RNN) naturally
arises. 

This framework provides us an insight on how the constructed RNN
be regularized.  For instance, one may regularize the model such
that the plus operator $\oplus$ is commutative. However, in this
paper, we do not explore further on this approach.

Note that this is different from \citep{Mikolov2013} where the
learned embeddings of words happened to be suitable for algebraic
operators. The operator-based framework proposed here is rather
geared toward \textit{learning} these operators directly.

\section{Experiments}
\label{sec:experiments}

We train four types of RNNs described in this paper on a number
of benchmark datasets to evaluate their performance. For each
benchmark dataset, we try the task of predicting the next symbol.

The task of predicting the next symbol is equivalent to the task
of modeling the distribution over a sequence.  For each sequence
$\left( \vx_1, \dots, \vx_T \right)$, we decompose it into 
\[
    p(\vx_1, \dots, \vx_T) = p(\vx_1) \prod_{t=2}^T p(\vx_t \mid \vx_1,
    \dots, \vx_{t-1}),
\]
and each term on the right-hand side will be replaced with a
single timestep of an RNN. In this setting, the RNN predicts the
probability of the next symbol $\vx_t$ in the sequence given the
all previous symbols $\vx_1,\dots\vx_{t-1}$. Then, we train the
RNN by maximizing the log-likelihood.

We try this task of modeling the joint distribution on three
different tasks; polyphonic music prediction, character-level and
word-level language modeling.

We test the RNNs on the task of polyphonic music prediction using
three datasets which are Nottingham, JSB Chorales and MuseData~\citep{Boulanger+al-ICML2012-small}. On the task of
character-level  and word-level language modeling, we use Penn
Treebank Corpus~\citep{Marcus1993}.

\subsection{Model Descriptions}

We compare the conventional recurrent neural network (RNN), deep
transition RNN with shortcut connections in the transition MLP
(DT(S)-RNN), deep output/transition RNN with shortcut connections
in the hidden to hidden transition MLP (DOT(S)-RNN) and stacked
RNN (sRNN).  See Fig.~\ref{fig:rnn_models} (a)--(d) for the
illustrations of these models. 

\begin{table}[ht]
    \centering
    \begin{tabular}{ c | c | c || c | c | c | c }
        \hline
        \multicolumn{3}{c||}{} & RNN & DT(S)-RNN & DOT(S)-RNN & sRNN  \\
        \multicolumn{3}{c||}{} & & & & 2 layers \\
        \hline 
        \hline
        \multirow{4}{*}{Music} 
        & Notthingam &\specialcell{\# units \\ \# parameters}
                    &\specialcell{600 \\465K} &
                       \specialcell{400,400 \\ 585K} & 
                       \specialcell{400,400,400 \\745K} &
                       \specialcell{ 400 \\ 550K}\\
        \hline
        & JSB Chorales  &\specialcell{\# units \\ \# parameters}
& \specialcell{200 \\ 75K} & 
                        \specialcell{400,400 \\ 585K } & 
                        \specialcell{400,400,400 \\745K} &
                        \specialcell{ 400 \\550K} \\
        \hline
        & MuseData  &\specialcell{\# units \\ \# parameters}
& \specialcell{600 \\ 465K} & 
                     \specialcell{400,400 \\585K} & 
                     \specialcell{400,400,400 \\ 745K} &
                     \specialcell{600 \\ 1185K} \\
        \hline
        \multirow{4}{*}{Language}
        & Char-level  &\specialcell{\# units \\ \# parameters}
& \specialcell{600 \\ 420K} & 
                       \specialcell{400,400 \\540K} & 
                       \specialcell{400,400,600 \\ 790K} &
                       \specialcell{400 \\ 520K} \\
        \hline
        & Word-level  &\specialcell{\# units \\ \# parameters}
& \specialcell{200 \\ 4.04M} &
                        \specialcell{200,200 \\6.12M} & 
                        \specialcell{200,200,200 \\ 6.16M} &
                        \specialcell{400 \\ 8.48M} \\
        \hline
    \end{tabular}
    \caption{The sizes of the trained models. We provide the number 
        of hidden units as well as the total number of parameters.
        For DT(S)-RNN, the two numbers provided for the number of units 
        mean the size of the
        hidden state and that of the intermediate layer,
        respectively. For DOT(S)-RNN, the three numbers are the size
        of the hidden state, that of the intermediate layer
        between the consecutive hidden states and that of the
        intermediate layer between the hidden state and the
        output layer. For sRNN, the number corresponds to the size of
the hidden state at each level}
    \label{tab:model_size}
\end{table}

The size of each model is chosen from a limited set $\left\{ 100,
200, 400, 600, 800 \right\}$ to minimize the validation error for
each polyphonic music task (See Table.~\ref{tab:model_size} for
the final models). In the case of language modeling tasks, we
chose the size of the models from $\left\{ 200, 400 \right\}$ and
$\left\{ 400, 600 \right\}$ for word-level and character-level
tasks, respectively.  In all cases, we use a logistic sigmoid
function as an element-wise nonlinearity of each hidden unit.
Only for the character-level language modeling we used rectified
linear units~\citep{Glorot+al-AI-2011-small} for the intermediate
layers of the output function, which gave lower validation error.

\subsection{Training}

We use stochastic gradient descent (SGD) and employ the strategy
of clipping the gradient proposed by
\citet{Pascanu+al-ICML2013-small}. Training stops when the
validation cost stops decreasing.

\textbf{Polyphonic Music Prediction}: 
For Nottingham and MuseData datasets we compute each gradient
step on subsequences of at most 200 steps, while we use
subsequences of 50 steps for JSB Chorales. We do not reset the
hidden state for each subsequence, unless the subsequence belongs
to a different song than the previous subsequence.

The cutoff threshold for the
gradients is set to 1. The hyperparameter for the
learning rate schedule\footnote{
    We use at each update $\tau$, the following learning rate
    $
        \eta_{\tau} = \frac{1}{1 + \frac{\max(0, \tau -
        \tau_0)}{\beta}},
        $
    where $\tau_0$ and $\beta$ indicate respectively when the learning rate
    starts decreasing and how quickly the learning rate decreases. In the
    experiment, we set $\tau_0$ to coincide with the time when the validation
    error starts increasing for the first time. 
} 
is tuned manually for each dataset.  We set the hyperparameter
$\beta$ to $2330$ for Nottingham, $1475$ for MuseData and $100$
for JSB Chroales.
They
correspond to two epochs, a single epoch and a third of an epoch, respectively.

The weights of the connections between any pair of hidden layers
are sparse, having only 20 non-zero incoming connections per unit~\citep[see, e.g.,][]{sutskeverimportance}. Each weight matrix is
rescaled to have a unit largest singular value~\citep{Pascanu+al-ICML2013-small}. The weights of the connections
between the input layer and the hidden state as well as between
the hidden state and the output layer are initialized randomly
from the white Gaussian distribution with its standard deviation
fixed to $0.1$ and $0.01$, respectively. In the case of deep
output functions (DOT(S)-RNN), the weights of the connections
between the hidden state and the intermediate layer are sampled
initially from the white Gaussian distribution of standard
deviation $0.01$. In all cases, the biases are initialized to
$0$.

To regularize the models, we add white Gaussian noise of standard
deviation $0.075$ to each weight parameter every time the
gradient is computed~\citep{Graves2011}.

\textbf{Language Modeling}: 
We used the same strategy for initializing the parameters in the
case of language modeling. For character-level modeling, the
standard deviations of the white Gaussian distributions for the
input-to-hidden weights and the hidden-to-output weights, we
used $0.01$ and $0.001$, respectively, while those hyperparameters 
were both $0.1$ for word-level modeling. In the case of DOT(S)-RNN, we sample the
weights of between the hidden state and the rectifier
intermediate layer of the output function from the white Gaussian
distribution of standard deviation $0.01$. When using rectifier
units (character-based language modeling) we fix the biases to
$0.1$.

In language modeling, the learning rate starts from an initial
value and is halved each time the validation cost does not
decrease significantly~\citep{Mikolov2010}. We do not use any
regularization for the character-level modeling, but for the
word-level modeling we use the same strategy of adding weight
noise as we do with the polyphonic music prediction.

For all the tasks (polyphonic music prediction, character-level
and word-level language modeling), the stacked RNN and the
DOT(S)-RNN were initialized with the weights of the conventional
RNN and the DT(S)-RNN, which is similar to layer-wise pretraining
of a feedforward neural network~\citep[see,
e.g.,][]{Hinton-Science2006}. We use a ten times smaller learning
rate for each parameter that was pretrained as either RNN or
DT(S)-RNN.

\begin{table}[ht]
    \centering
    \begin{tabular}{c | c c c c | c}
        \hline
        & RNN & DT(S)-RNN & DOT(S)-RNN & sRNN & DOT(S)-RNN* \\
        \hline
        \hline
        Notthingam & $3.225$ & $3.206$ & $3.215$ & $3.258$ & $2.95$ \\
        JSB Chorales & $8.338$ & $8.278$ & $8.437$ & $8.367$ & $7.92$ \\
        MuseData & $6.990$ & $6.988$ & $6.973$ & $6.954$ & $6.59$ \\
        \hline
    \end{tabular}
    \caption{The performances of the four types of RNNs on the
        polyphonic music prediction. The numbers represent
        negative log-probabilities on test sequences. (*) We
        obtained these results using DOT(S)-RNN with $L_p$ units
        in the deep transition, maxout units in the deep output
        function and dropout~\citep{Gulcehre2013}.}
    \label{tab:result_music}
\end{table}

\subsection{Result and Analysis}

\subsubsection{Polyphonic Music Prediction}

The log-probabilities on the test set of each data are presented
in the first four columns of Tab.~\ref{tab:result_music}. We were
able to observe that in all cases one of the proposed deep RNNs
outperformed the conventional, shallow RNN. Though, the
suitability of each deep RNN depended on the data it was trained
on.  The best results obtained by the DT(S)-RNNs on Notthingam
and JSB Chorales are close to, but worse than the result obtained
by RNNs trained with the technique of fast dropout (FD) which are
$3.09$ and $8.01$, respectively~\citep{Justin2013}. 

In order to quickly investigate whether the proposed deeper
variants of RNNs may also benefit from the recent advances in
feedforward neural networks, such as the use of non-saturating
activation functions\footnote{
    Note that it is not trivial to use non-saturating activation
    functions in conventional RNNs, as this may cause the
    explosion of the activations of hidden states. However, it is
    perfectly safe to use non-saturating activation functions at
    the intermediate layers of a deep RNN with deep transition.
}
and the method of dropout.  We have built another set of DOT(S)-RNNs that have
the recently proposed $L_p$ units~\citep{Gulcehre2013} in deep transition and
maxout units~\citep{Goodfellow_maxout_2013} in deep output function.
Furthermore, we used the method of dropout~\citep{Hinton-et-al-arxiv2012}
instead of weight noise during training. Similarly to the previously trained
models, we searched for the size of the models as well as other learning
hyperparameters that minimize the validation performance. We, however, did not
pretrain these models.

The results obtained by the DOT(S)-RNNs having $L_p$ and maxout
units trained with dropout are shown in the last column of
Tab.~\ref{tab:result_music}. On every music dataset the
performance by this model is significantly better than those
achieved by all the other models as well as the best results
reported with recurrent neural networks in \citep{Justin2013}.
This suggests us that the proposed variants of deep RNNs also
benefit from having non-saturating activations and using dropout,
just like feedforward neural networks. We reported these results
and more details on the experiment in \citep{Gulcehre2013}.

We, however, acknowledge that the model-free state-of-the-art
results for the both datasets were obtained using an RNN combined
with a conditional generative model, such as restricted Boltzmann
machines or neural autoregressive distribution
estimator~\citep[][]{Larochelle+Murray-2011}, in the
output~\citep{Boulanger+al-ICML2012-small}.  

\begin{table}[htb]
    \centering
    \begin{tabular}{c | c c c c | c c }
        \hline
        & RNN & DT(S)-RNN & DOT(S)-RNN & sRNN & $*$ & $\star$ \\
        \hline
        \hline
        Character-Level & $1.414$ & $1.409$ & $\mathbf{1.386}$ & $1.412$ & 
        $1.41$\footnotemark[1] & $1.24$\footnotemark[3] \\
        Word-Level & $117.7$ & $112.0$ & $\mathbf{107.5}$ & $110.0$ & 
        $123$\footnotemark[2] & $117$\footnotemark[3] \\
        \hline
    \end{tabular}
    \caption{The performances of the four types of RNNs on the
        tasks of language modeling. The numbers represent
        bit-per-character and perplexity computed on test
        sequence, respectively, for the character-level and
        word-level modeling tasks. $*$ The previous/current
    state-of-the-art results obtained with shallow RNNs. $\star$
The previous/current state-of-the-art results obtained with
RNNs having long-short term memory units.}
    \label{tab:result_lm_penn}
\end{table}

\footnotetext[1]{Reported by \citet{TomasIlya} using mRNN with 
Hessian-free optimization technique.}
\footnotetext[2]{Reported by \citet{Mikolov-ICASSP-2011} using
the dynamic evaluation.}
\footnotetext[3]{Reported by \citet{Graves2013} using the dynamic
evaluation and weight noise.}

\subsubsection{Language Modeling}

On Tab.~\ref{tab:result_lm_penn}, we can see the perplexities on
the test set achieved by the all four models. We can clearly see
that the deep RNNs (DT(S)-RNN, DOT(S)-RNN and sRNN) outperform
the conventional, shallow RNN significantly.  On these tasks
DOT(S)-RNN outperformed all the other models, which suggests that
it is important to have highly nonlinear mapping from the hidden
state to the output in the case of language modeling.

The results by both the DOT(S)-RNN and the sRNN for word-level
modeling surpassed the previous best performance achieved by an
RNN with 1000 long short-term memory (LSTM) units~\citep{Graves2013} as well as that by a shallow RNN with a larger
hidden state~\citep{Mikolov-ICASSP-2011}, even when both of them
used dynamic evaluation\footnote{Dynamic evaluation refers to an
    approach where the parameters of a model are updated as the
validation/test data is predicted.}. The results we report here are without
dynamic evaluation.

For character-level modeling the state-of-the-art results were
obtained using an optimization method Hessian-free with a specific type of
RNN architecture called mRNN~\citep{TomasIlya} or a regularization
technique called adaptive weight noise~\citep{Graves2013}. Our result, however, 
is better than the performance achieved by conventional, 
shallow RNNs without any of those advanced regularization methods~\citep{Mikolov-Sutskever-2012}, where they reported the best
performance of $1.41$ using an RNN trained with the Hessian-free
learning algorithm~\citep{Martens+Sutskever-ICML2011}.

\section{Discussion}
\label{sec:final}

In this paper, we have explored a novel approach to building a
deep recurrent neural network (RNN). We considered the structure
of an RNN at each timestep, which revealed that the relationship
between the consecutive hidden states and that between the hidden
state and output are \textit{shallow}. Based on this observation,
we proposed two alternative designs of \textit{deep} RNN that
make those \textit{shallow} relationships be modeled by deep
neural networks. Furthermore, we proposed to make use of shortcut
connections in these deep RNNs to alleviate a problem of
difficult learning potentially introduced by the increasing
depth.

We empirically evaluated the proposed designs against the
conventional RNN which has only a single hidden layer and against
another approach of building a deep RNN~\citep[stacked
RNN,][]{Graves2013}, on the task of polyphonic music prediction
and language modeling. 

The experiments revealed that the RNN with the proposed deep
transition and deep output (DOT(S)-RNN) outperformed both the
conventional RNN and the stacked RNN on the task of language
modeling, achieving the state-of-the-art result on the task of
word-level language modeling. For polyphonic music prediction, a
different deeper variant of an RNN achieved the best performance
for each dataset. Importantly, however, in all the cases, the
conventional, shallow RNN was not able to outperform the deeper
variants.  These results strongly support our claim that an RNN
benefits from having a deeper architecture, just like feedforward
neural networks.

The observation that there is no clear winner in the task of
polyphonic music prediction suggests us that each of the proposed
deep RNNs has a distinct characteristic that makes it more, or
less, suitable for certain types of datasets. We suspect that in
the future it will be possible to design and train yet another
deeper variant of an RNN that combines the proposed models
together to be more robust to the characteristics of datasets.
For instance, a stacked DT(S)-RNN may be constructed by combining
the DT(S)-RNN and the sRNN.

In a quick additional experiment where we have trained DOT(S)-RNN
constructed using non-saturating nonlinear activation functions
and trained with the method of dropout, we were able to improve
the performance of the deep recurrent neural networks on the
polyphonic music prediction tasks significantly. This suggests us
that it is important to investigate the possibility of applying
recent advances in feedforward neural networks, such as novel,
non-saturating activation functions and the method of dropout, to
recurrent neural networks as well. However, we leave this as
future research.

One practical issue we ran into during the experiments was the
difficulty of training deep RNNs. We were able to train the
conventional RNN as well as the DT(S)-RNN easily, but it was not
trivial to train the DOT(S)-RNN and the stacked RNN. In this
paper, we proposed to use shortcut connections as well as to
pretrain them either with the conventional RNN or with the
DT(S)-RNN. We, however, believe that learning may become even
more problematic as the size and the depth of a model increase.
In the future, it will be important to investigate the root
causes of this difficulty and to explore potential solutions.  We
find some of the recently introduced approaches, such as advanced
regularization methods~\citep{Pascanu+al-ICML2013-small} and
advanced optimization algorithms~\citep[see,
e.g.,][]{Pascanu+Bengio-arxiv2013,martens2010hessian}, to be
promising candidates.

{
\subsubsection*{Acknowledgments}
We would like to thank the developers of
Theano~\citep{bergstra+al:2010-scipy,Bastien-Theano-2012}.  We
also thank Justin Bayer for his insightful comments on the paper.
We would like to thank NSERC, Compute Canada, and Calcul Qu\'ebec
for providing computational resources. Razvan Pascanu is
supported by a DeepMind Fellowship.  Kyunghyun Cho is supported
by FICS (Finnish Doctoral Programme in Computational Sciences)
and ``the Academy of Finland (Finnish Centre of Excellence in
Computational Inference Research COIN, 251170)''.

\newpage
\bibliography{strings,strings-short,strings-shorter,aigaion-shorter,ml,myref}
\bibliographystyle{natbib}
}

\end{document}